\documentclass[11pt]{article}
\usepackage[final]{acl}
\usepackage{times}
\usepackage{latexsym}
\usepackage[T1]{fontenc}
\usepackage[utf8]{inputenc}
\usepackage{microtype}
\usepackage{inconsolata}
\usepackage{graphicx}
\usepackage{todonotes}
\usepackage{comment}
\usepackage{booktabs}
\usepackage{multirow}
\usepackage{subcaption}
\usepackage{amssymb}
\usepackage{amsmath}
\usepackage{mathtools}
\usepackage[most]{tcolorbox}
\usepackage{xcolor}
\usepackage{fancyvrb}
\usepackage{fvextra}
\usepackage{url}
\usepackage{caption}
\usepackage{wrapfig}

\definecolor{promptbg}{RGB}{248,249,252}
\definecolor{promptframe}{RGB}{70,90,140}
\definecolor{darkgreen}{RGB}{0,100,0}
\definecolor{lightgray}{RGB}{240,240,240}

\newtcolorbox{promptbox}[1]{
  enhanced, breakable,
  colback=promptbg,
  colframe=promptframe,
  coltitle=white,
  fonttitle=\bfseries\small,
  title=#1,
  boxrule=0.4pt,
  left=4pt, right=4pt, top=3pt, bottom=3pt,
  arc=2pt,
}

\title{TeacherGRPO: Closing the Capacity Gap in Reasoning Distillation via Teacher Alignment}

\author{
    Zhenyu Lei$^{\blacklozenge}$ \:
    Zihan Chen$^\blacklozenge$ \:
    Yaochen Zhu$^\clubsuit$ \:
    Shangbin Feng$^\spadesuit$ \\ \bf
    Zaiyi Zheng$^\blacklozenge$ \:
    Ruocheng Guo$^\dagger$ \:
    Yushun Dong$^\heartsuit$ \:
    Jundong Li$^\blacklozenge$ \\
    $^\blacklozenge$University of Virginia, $^\clubsuit$Netflix, $^\spadesuit$University of Washington, $^\heartsuit$Florida State University, $^\dagger$Microsoft\\
    \texttt{\{vjd5zr, brf3rx, sjc4fq, jundong\}@virginia.edu}\\
    \texttt{yzhu@netflix.com, shangbin@cs.washington.edu, yd24f@fsu.edu, rguo.asu@gmail.com}
}

\begin{document}
\maketitle

\begin{abstract}
Reasoning distillation from powerful teacher models to smaller students faces the Gap Curse: as teachers grow more sophisticated, their complex distributions increasingly diverge from what students can approximate, causing performance degradation. Existing mitigation strategies either filter out challenging examples through data selection or introduce weaker intermediate assistant models, inherently compromising supervision coverage or quality. We propose Teacher Alignment, which directly adapts the teacher toward the student's distribution without discarding data or degrading reasoning quality. However, naive alignment through standard knowledge distillation triggers catastrophic collapse of the teacher's reasoning capabilities. To address this, we reformulate teacher alignment as reinforcement learning and introduce TeacherGRPO, built on Group Relative Policy Optimization with two key innovations: (i) Curriculum Selective Alignment applies dual token- and distribution-level curricula to focus rewards on high-signal reasoning gaps while filtering noise from trivial tokens and uncertain tail distributions, and (ii) Importance-Adaptive Length Regularization selectively penalizes verbose redundancy while preserving pedagogically critical reasoning steps. The aligned teacher then distills knowledge to students via standard pipelines. Extensive experiments show TeacherGRPO significantly outperforms baselines across diverse reasoning benchmarks and distillation methods. Our code is available at \url{https://github.com/LzyFischer/TeacherGRPO}.
\end{abstract}

\section{Introduction}

\begin{figure}[t]
  \centering
  \includegraphics[width=\linewidth]{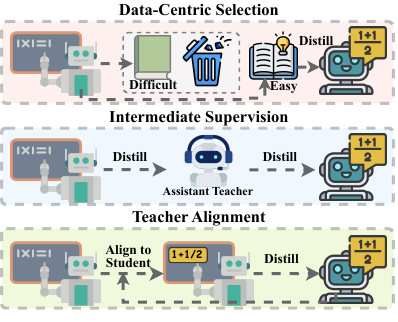}
  \caption{\textbf{Approaches to address the Gap Curse.} Data-Centric Selection discards potentially valuable examples while Intermediate Supervision dilutes supervision quality. Teacher Alignment adapts the teacher to student capacity, preserving both data and teaching quality.}
  \label{fig:reliability}
\end{figure}

Large Language Models (LLMs) have achieved remarkable reasoning milestones~\citep{xu2025towards}, yet their computational cost remains a significant barrier for resource-constrained deployment~\citep{bai2024beyond}. While reasoning distillation offers a promising path to transfer these capabilities to smaller models~\citep{kang2023knowledge, fu2023specializing}, it faces a counterintuitive hurdle known as the ``Gap Curse''~\citep{zhang2025towards, zhu2022teach, jiang2025drp, zhanmakes}. As teacher models become increasingly sophisticated, their complex reasoning patterns diverge distributionally from what capacity-limited students can approximate, which leads to degraded student performance.
Current approaches to mitigating this gap fall into two primary paradigms. The first employs \textit{Data-Centric Selection} strategies, utilizing difficulty-aware metrics to filter samples with excessive teacher-student divergence~\citep{liu2025less, huang2025selectkd, li2025sample}. However, these methods often discard valuable challenging examples that could benefit learning. The second paradigm focuses on \textit{Intermediate Supervision}, introducing intermediate teacher assistant models to bridge the complexity gap between teacher and student~\citep{zhou2024teaching, lv2024taekd, ding2025micota}. However, these assistants typically exhibit inferior reasoning capabilities compared to the original teacher, potentially propagating errors or suboptimal reasoning patterns to the student~\citep{budiman2025survey}.

As both paradigms inherently compromise supervision quality, recent knowledge distillation work suggests a third path~\citep{qian2025good}: adapting the teacher toward the student's distribution via standard KD~\citep{hinton2015distilling}, which we term \textit{Teacher Alignment}. By shifting the teacher's distribution closer to the student without discarding high-quality data or replacing the original teacher with a weaker surrogate, this paradigm produces more accessible supervision signals while preserving the supervision quality at the source. However, naively applying Teacher Alignment to reasoning distillation introduces a critical challenge: reasoning ability is fragile and uniquely sensitive to distribution shift, making the teacher prone to collapse under aggressive fine-tuning~\citep{li2024continual} -- paradoxically \textit{worsening} the supervision quality it was designed to protect. Our preliminary experiments (Section~\ref{sec:preliminary}) confirm this fatal flaw: as the teacher is aligned toward the student's distribution, its reasoning capabilities rapidly deteriorate, rendering it an even less effective knowledge source.

Inspired by recent findings demonstrating that reinforcement learning (RL) can mitigate model collapse in LLMs~\citep{kumar2024training, korbak2022reinforcement}, we reformulate Teacher Alignment as an RL problem. Unlike direct distillation, which forces rigid distribution mimicry, our approach treats the student's capacity as a reward signal. This allows the teacher to actively explore and discover reasoning paths that are digestible for the student, while maintaining its own logical correctness~\citep{yue2025does}. However, directly applying RL to Teacher Alignment introduces two specific challenges:

\noindent\textbf{(1) Diluted Alignment Signals.} Standard KD rewards obscure effective supervision. (i) \textit{Token-level}: The reward objective indiscriminately incentivizes the matching of trivial tokens (e.g. ``The answer is'') that are irrelevant to reasoning capability~\citep{ye2025disentangling, gao2024designing}, preventing the teacher from understanding the genuine reasoning gaps where the student struggles. (ii) \textit{Distribution-level}: The reward is contaminated by the long tail regions where both models are uncertain~\citep{zhai2023uncertainty}. Considering rewards from these irrelevant low-probability tokens injects noise rather than meaningful knowledge gap.

\noindent\textbf{(2) Length Exploitation.} Capacity-constrained students are prone to error propagation in long chains~\citep{ji2025chain}, necessitating concise reasoning~\citep{jiang2025teach}. However, naive alignment creates perverse incentives for verbosity: teachers can trivially minimize divergence by generating repetitive, low-information chains that mirror the student's uncertainty~\citep{wang2025thinking}. This effectively hacks the reward without improving pedagogical quality.

To address these challenges, we introduce TeacherGRPO, a framework based on Group Relative Policy Optimization~\citep{shao2024deepseekmath} with two core innovations:
First, to eliminate reward dilution, we propose \textbf{Curriculum Selective Alignment (CSA)}. Rather than optimizing against all tokens and full vocabulary, we employ a dual curriculum strategy: (i) \textit{token-level curriculum} initially targeting the top-$k$ tokens with highest teacher-student gaps instead of trivial ones, and (ii) \textit{distribution-level curriculum} beginning with the top-$K$ probability-divergence entries that capture core distributional differences, then gradually expanding coverage. This hierarchical approach filters out statistical noise, focusing reward solely on high-signal reasoning gaps.
Second, to prevent length exploitation, we design \textbf{Importance-Adaptive Length Regularization (IALR)} to penalize over verbose reasoning paths. Unlike uniform penalties that indiscriminately penalize all steps that risks truncating critical reasoning steps~\citep{jin2024impact}, IALR scales penalty by each step's estimated importance. This selectively suppresses redundant reasoning steps while preserving pedagogically essential reasoning, enabling more effective knowledge transfer.

After Teacher Alignment with TeacherGRPO, we employ the optimized teacher to distill knowledge into student models using standard distillation techniques. Extensive experiments demonstrate that TeacherGRPO provides substantially superior supervision signals compared to both vanilla and KD-aligned teachers, yielding significant improvements in student performance across diverse reasoning benchmarks.

\begin{figure*}[t]
    \centering
    \includegraphics[width=\linewidth]{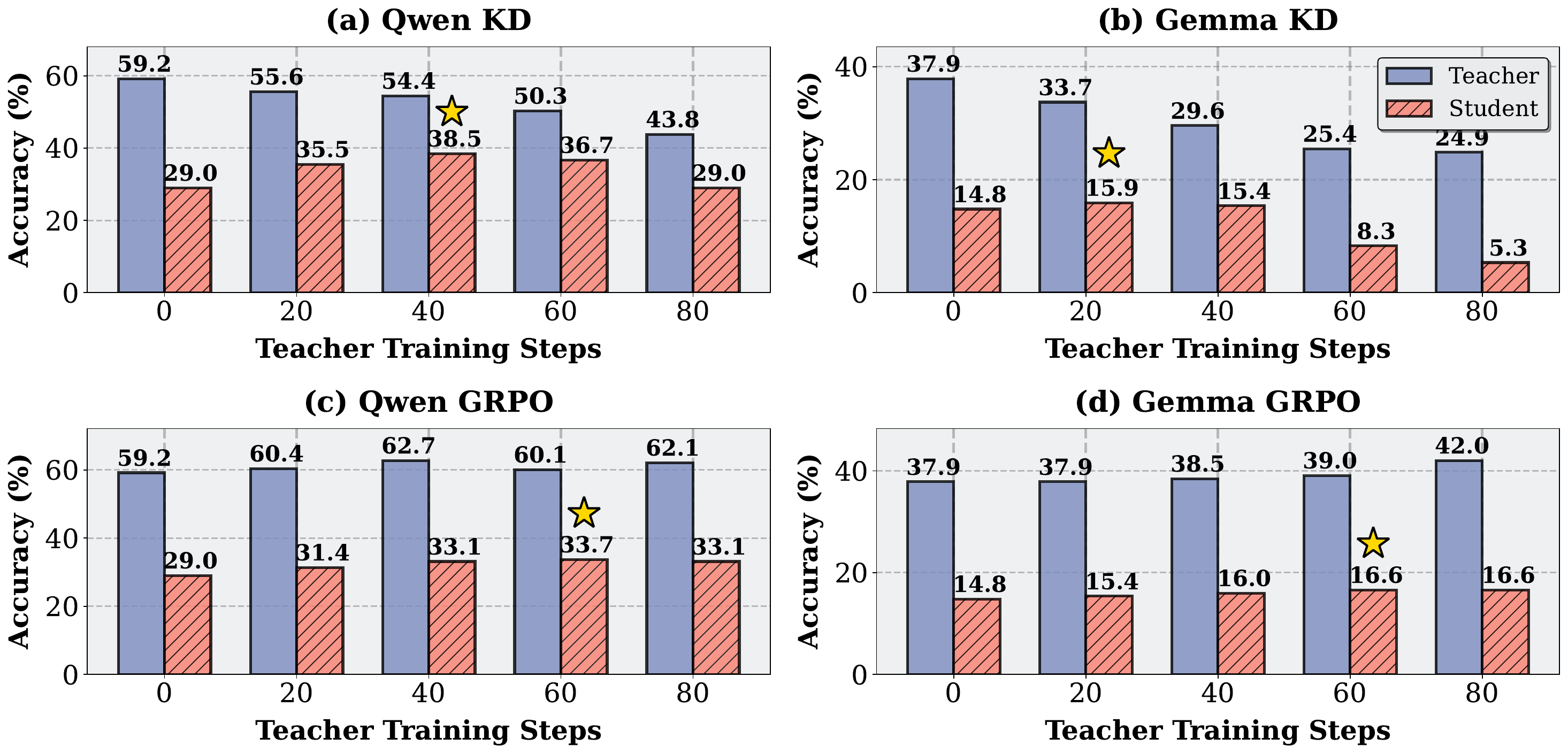}
    \caption{\textbf{Teacher and student performance during alignment.} Stars indicate peak student performance. KD-based alignment initially improves distillation but causes catastrophic teacher collapse. GRPO maintains teacher capability but achieves limited alignment effectiveness, motivating TeacherGRPO.}
    \label{fig:preliminary}
\end{figure*}

\section{Preliminary Study}
\label{sec:preliminary}
To motivate our approach, we investigate the feasibility and limitations of \textit{Teacher Alignment}. Specifically, we aim to answer two questions: (1) whether aligning the teacher to the student's distribution improves downstream distillation, and (2) whether naive alignment strategies are sufficient.

\noindent\textbf{Experimental Setup.} We compare two alignment methods on teacher models (\texttt{Qwen2.5-3B-Instruct}~\citep{yang2025qwen3} and \texttt{Gemma-3-1b-it}~\citep{team2025gemma}) to align them with their student counterparts (\texttt{Qwen2.5-0.5B-Instruct} and \texttt{Gemma} \texttt{-3-270m-it}): (1) \textbf{KD}: minimizes KL divergence via logit-based knowledge distillation, and (2) \textbf{GRPO}: optimizes teacher distribution using GRPO with reverse KL divergence as rewards. Both methods train on Date Understanding dataset~\citep{srivastava2023beyond} for 80 steps (checkpoints at steps 0, 20, 40, 60, 80) with learning rate 5e-6 and 2e-5. We evaluate: (1) \textbf{Teacher Capability}---teacher validation accuracy, and (2) \textbf{Distillation Efficacy}---accuracy of students distilled from each teacher checkpoint.

\noindent\textbf{Observation 1: Teacher Alignment Improves Distillation.} Figure~\ref{fig:preliminary} shows that both KD and GRPO reduce teacher-student divergence and improve distillation. Specifically, Qwen student accuracy increases from 30.2\% (baseline) to 38.5\% (KD, step 40) and 33.1\% (GRPO, step 80), confirming that teachers aligned to students' distributional characteristics provide better supervision.

\noindent\textbf{Observation 2: KD Causes Catastrophic Capability Collapse.} Beyond the optimal checkpoint (marked with stars), KD-aligned teachers suffer severe performance degradation. Specifically, Qwen teacher accuracy drops from 59.2\% to 43.8\% by step 80, while student's plummets from peak 38.5\% to 29.0\%. By forcing exact distribution mimicry, KD causes the teacher to overfit to the student's errors and inferior reasoning patterns, corrupting it as a knowledge source and degrading late-stage distillation performance.

\noindent\textbf{Observation 3: GRPO Maintains Capability but Limits Alignment Effectiveness.} GRPO-aligned teachers maintain stable reasoning capabilities throughout training. Qwen teacher accuracy remains at 58.5\% at step 80, and Gemma maintains 36.2\%. By treating alignment as reward optimization rather than distribution mimicry, GRPO explores reasoning paths accessible to students without sacrificing logical correctness. However, GRPO-distilled students achieve lower peak performance than KD. This limitation motivates TeacherGRPO to enhance GRPO's alignment effectiveness while preserving capability stability.

\noindent\textbf{Conclusion.} This study reveals a critical dilemma: we must bridge the distributional gap while preserving teacher reasoning capabilities. KD achieves strong alignment but causes catastrophic capability degradation. GRPO maintains capability stability but achieves insufficient alignment effectiveness. This motivates TeacherGRPO to enhance GRPO's alignment while preserving its capability stability.

\section{Methodology}
\label{sec:methodology}

\begin{figure*}[t]
    \centering
    \includegraphics[width=0.9\linewidth]{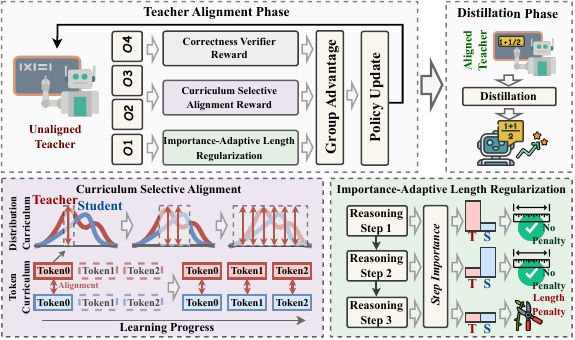}
    \caption{\textbf{TeacherGRPO framework overview.} The Alignment phase optimizes the teacher via GRPO with two novel reward CSA and IALR, plus a correctness verification. The aligned teacher then supervises student distillation.}
    \label{fig:overview}
\end{figure*}

We propose \textbf{TeacherGRPO}, a reinforcement learning framework that aligns teacher reasoning capabilities with student capacity while preserving the teacher's superior reasoning abilities. Our approach addresses two fundamental challenges in GRPO-based Teacher Alignment: (1) diluted alignment signals from token-level and distribution-level noise, and (2) length exploitation that produces verbose, pedagogically ineffective outputs. We tackle these through two novel reward mechanisms: \textit{Curriculum Selective Alignment (CSA)} and \textit{Importance-Adaptive Length Regularization (IALR)}.

\subsection{Problem Formulation}

Given a teacher model with parameters $\theta$ generating outputs according to policy $\pi_\theta$, and a student model with fixed policy $\pi_S$, our objective is to adapt $\pi_\theta$ to align with the reasoning ability of $\pi_S$ within GRPO. Formally, we seek to optimize:
\begin{equation*}
\begin{split}
    \theta^* = \arg\max_{\theta}\; & \mathbb{E}_{x \sim \mathcal{D},\, y \sim \pi_\theta(\cdot|x)} \left[ R_{align}(y; \pi_\theta, \pi_S) \right] \\
    & - \lambda D_{KL}(\pi_\theta \,\|\, \pi_{ref})
\end{split}
\end{equation*}
where $\mathcal{D}$ denotes the dataset and $D_{KL}$ denotes the KL penalty against reference policy $\pi_{ref}$ that prevents capability collapse. The core innovation of TeacherGRPO lies in the design of $R_{align}$, which we decompose into CSA and IALR components.

\subsection{Curriculum Selective Alignment (CSA)}
\label{subsec:csa}
Standard logit-based reward often suffer from signal dilution due to two sources of noise: \textit{token-level noise}, arising from trivial tokens that obscure genuine reasoning gap, and \textit{distribution-level noise}, stemming from non-informative probability mass in the vocabulary tail. To address this, we introduce Curriculum Selective Alignment, a dual-curriculum strategy that progressively filters these noise to focus on genuine reasoning discrepancies.

\subsubsection{Token-Level Curriculum}

Building on prior research indicating that distributional divergence often signal underlying reasoning disparities~\citep{wang2025lightreasoner}, we focus the alignment on critical tokens with high KL divergence. For each token position $t$, we compute $D_t = D_{KL}(\pi_\theta(\cdot|y_{<t}) \| \pi_S(\cdot|y_{<t}))$ where $y_{<t}$ denotes the teacher-generated history. To prioritize learning on these pivots, we generate a binary selection mask $M_t$, defined as $M_t = 1$ if the rank of $D_t$ falls within the top $k$ percentile of the sequence, and $0$ otherwise. However, a static $k$ is suboptimal, as it risks overlooking subtle reasoning gaps that become relevant as alignment improves. Therefore, we implement a dynamic curriculum where the selection threshold $k(\tau)$ expands linearly over training steps $\tau$:
\begin{equation}
k(\tau) = k_{\text{min}} + (k_{\text{max}} - k_{\text{min}}) \cdot \frac{\tau}{T}
\end{equation}
where $T$ represents the total training steps. This progression ensures the teacher resolves fundamental reasoning disconnects (high $D_t$) before gradually fine-tuning on more subtle reasoning logic.

\subsubsection{Distribution-Level Curriculum}
Beyond token selection, the alignment signal is susceptible to noise from the vocabulary tail, where both models are uncertain and assign small probability. To mitigate high-variance gradients from these regions, we constrain the divergence calculation to
$\mathcal{V}_p(y_t) = \{ w \in \mathcal{V} \mid \text{rank}(|\pi_\theta(w|y_{<t}) - \pi_S(w|y_{<t})|) \le p \}$, the subset of the top-$p$ entries with highest probability gap. Combined with token-level mask $M_t$, the final truncated alignment reward at position $t$ is defined as:
\begin{equation}
\resizebox{\columnwidth}{!}{$
R_{\text{vocab}}(y_t; p) = - \sum_{w \in \mathcal{V}_p(y_t)} \pi_\theta(w | y_{<t}) \log \left( \frac{\pi_\theta(w | y_{<t})}{\pi_S(w | y_{<t})} \right) * M
$}
\end{equation}
Analogous to the token-level strategy, we implement a progressive curriculum for the vocabulary coverage $p(\tau)$. This threshold linearly interpolates from a focused core to the full vocabulary $|\mathcal{V}|$ over the training course $\tau$. This schedule ensures the teacher initially aligns with the student's dominant probability modes before adapting to the subtle uncertainties in the distributional tail.

\subsection{Importance-Adaptive Length Regularization}

Teacher models often exhibit a tendency toward verbosity. A common countermeasure is a uniform length penalty in the reward function, but this can inadvertently penalize essential reasoning steps necessary for effective student comprehension. To address this limitation, we propose \textbf{Importance-Adaptive Length Regularization (IALR)}, which selectively penalizes low-information verbosity while preserving critical reasoning pivots.

\subsubsection{Step Importance Quantification}
To identify and preserve critical steps, we firstly decompose the reasoning chain $y$ into logical steps $S = \{s_1, s_2, \dots, s_N\}$ based on structural delimiters (e.g., newlines).
In reasoning tasks, step importance is commonly quantified through predictive uncertainty~\citep{zeng2025pruning}: high-uncertainty steps represent critical decision points requiring careful reasoning, while low-uncertainty steps indicate routine continuations. In addition, for teacher alignment, a step's pedagogical value depends on both teacher and student: a step is essential if it represents (i) a critical reasoning pivot for the teacher, preserving logical coherence, or (ii) a challenging transition where the student must follow complex reasoning.
Therefore, we define the \emph{joint importance score} for step $s_i$ as
\begin{equation}
\resizebox{0.85\linewidth}{!}{$
\mathcal{I}_i = \sigma\!\left(\frac{\mathcal{H}\bigl(\pi_\theta(\cdot \mid y_{<t_i})\bigr) + \mathcal{H}\bigl(\pi_S(\cdot \mid y_{<t_i})\bigr)}{\eta}\right)
$}
\end{equation}
where $\mathcal{H}(\pi) = -\sum_w \pi(w) \log \pi(w)$ denotes entropy, $t_i$ denotes the token index marking the start of step $s_i$, $\sigma$ is the sigmoid function, and $\tau$ controls sensitivity. High joint entropy indicates pivotal reasoning points, whereas low entropy suggests routine elaboration.

\subsubsection{Importance-Weighted Length Penalty}

We then apply a length penalty that is inversely proportional to semantic importance, producing the IALR reward:
\[
R_{\text{IALR}}(y) = -\sum_{i=1}^{N} \beta\,|s_i|\,\bigl(1 - \mathcal{I}_i\bigr)\,
\]
where $|s_i|$ denotes the token count of step $s_i$ and $\beta$ controls the penalty's strength. This formulation penalizes long, low-entropy steps more heavily while imposing minimal penalties on concise or high-entropy reasoning, thereby yielding concise yet complete reasoning chains optimized for student learning.

\begin{table*}[t]
\centering
\caption{Performance comparison across four reasoning benchmarks with two model families: Qwen and Gemma. Best average values are in bold.}
\label{tab:main_results}
\resizebox{0.85\textwidth}{!}{
\begin{tabular}{lccccc|ccccc}
\toprule
\multirow{3}{*}{\textbf{Method}} & \multicolumn{5}{c|}{\textbf{Qwen2.5 ($\textbf{3B} \to \textbf{0.5B}$)}} & \multicolumn{5}{c}{\textbf{Gemma3 ($\textbf{1B} \to \textbf{270M}$)}} \\
\cmidrule(lr){2-6} \cmidrule(lr){7-11}
& \textbf{Date} & \textbf{SQA} & \textbf{ARC} & \textbf{CQA} & \textbf{Avg} & \textbf{Date} & \textbf{SQA} & \textbf{ARC} & \textbf{CQA} & \textbf{Avg} \\
\midrule
\multicolumn{11}{l}{\textit{\textbf{Teacher}}} \\
\midrule
Base & 58.6 & 59.8 & 81.7 & 71.5 & 67.9 & 37.9 & 51.1 & 47.9 & 47.3 & 46.1 \\
\hspace{3mm}\textit{+ TeacherKD} & 53.9 & 55.5 & 79.2 & 61.2 & 62.5 & 31.5 & 47.2 & 46.3 & 44.3 & 43.8 \\
\hspace{3mm}\textit{+ \textbf{TeacherGRPO}} & 59.2 & 64.6 & 81.2 & 72.7 & \textbf{69.4} & 40.2 & 51.5 & 48.3 & 47.1 & \textbf{46.8} \\
\midrule
\multicolumn{11}{l}{\textit{\textbf{Student}}} \\
\midrule
Base & 20.1 & 35.8 & 37.4 & 39.1 & 33.1 & 13.6 & 28.8 & 12.5 & 13.9 & 17.2 \\
\cmidrule(lr){2-6} \cmidrule(lr){7-11}
KD & 29.0 & 39.7 & 40.7 & 47.3 & 39.2 & 14.8 & 35.4 & 24.7 & 21.5 & 24.1 \\
\hspace{3mm}\textit{+ TeacherKD} & 38.5 & 47.6 & 45.4 & 26.8 & 39.6 & 15.4 & 38.4 & 24.1 & 21.0 & 24.7 \\
\hspace{3mm}\textit{+ \textbf{TeacherGRPO}} & 40.2 & 44.5 & 45.2 & 51.9 & \textbf{45.5} & 17.8 & 42.8 & 25.7 & 22.1 & \textbf{27.1} \\
\cmidrule(lr){2-6} \cmidrule(lr){7-11}
SeqKD & 34.6 & 38.4 & 36.4 & 40.3 & 37.4 & 15.4 & 32.8 & 23.1 & 20.5 & 23.0 \\
\hspace{3mm}\textit{+ TeacherKD} & 38.5 & 43.3 & 34.1 & 41.4 & 39.3 & 14.8 & 36.2 & 23.4 & 20.7 & 23.8 \\
\hspace{3mm}\textit{+ \textbf{TeacherGRPO}} & 38.8 & 42.7 & 41.3 & 41.7 & \textbf{41.1} & 17.8 & 36.2 & 25.4 & 23.1 & \textbf{25.6} \\
\cmidrule(lr){2-6} \cmidrule(lr){7-11}
Adaptive Off-Policy & 20.7 & 38.0 & 41.9 & 46.4 & 36.8 & 16.6 & 40.2 & 22.3 & 16.8 & 24.0 \\
\hspace{3mm}\textit{+ TeacherKD} & 26.0 & 39.7 & 38.1 & 28.3 & 33.0 & 8.9 & 47.2 & 19.9 & 19.7 & 23.9 \\
\hspace{3mm}\textit{+ \textbf{TeacherGRPO}} & 30.2 & 42.4 & 44.6 & 48.3 & \textbf{41.4} & 20.7 & 46.3 & 23.9 & 18.8 & \textbf{27.4} \\
\bottomrule
\end{tabular}
}
\end{table*}

\subsection{TeacherGRPO Optimization}
\label{subsec:teacher_grpo}

We integrate CSA and IALR into GRPO for stable and effective alignment. For each query $x$, we sample $G$ outputs $\{y_1, \dots, y_G\}$ from policy $\pi_{\theta_{old}}$. The total reward combines:
\begin{equation*}
\resizebox{\linewidth}{!}{$
    R_{total}(y_j) = \alpha_1 R_{CSA}(y_j) + \alpha_2 R_{SALR}(y_j) + \alpha_3 R_{ver}(y_j)
$}
\end{equation*}
where $R_{ver}(y_j) \in \{0, 1\}$ is a correctness verifier, and $\{\alpha_1, \alpha_2, \alpha_3\}$ balance alignment, compression, and correctness. We compute group-normalized advantages and optimize:
\begin{equation}
\resizebox{0.85\linewidth}{!}{$
\displaystyle
\begin{aligned}
    \mathcal{L}(\theta) = & -\frac{1}{G} \sum_{j=1}^{G} \Bigg[ \frac{1}{|y_j|} \sum_{t=1}^{|y_j|} \min \Big( r_t(\theta) A_j, \text{clip}(r_t(\theta), \\ & 1-\epsilon_{clip},  1+\epsilon_{clip}) A_j \Big) \Bigg]
     + \gamma D_{KL}(\pi_\theta || \pi_{ref})
\end{aligned}
$}
\end{equation}
where $r_t(\theta) = \frac{\pi_\theta(y_{j,t}|y_{j,<t})}{\pi_{\theta_{old}}(y_{j,t}|y_{j,<t})}$, $A_j = \frac{R_{total}(y_j) - \mu_G}{\sigma_G + \epsilon}$ is the advantage normalized by the mean $\mu_G$ and standard deviation $\sigma_G$ of $R_{total}$ with a small $\epsilon>0$ for numerical stability, and $\pi_{ref}$ is the original teacher. The clip prevents destructive updates, while the KL preserves teacher capabilities.

Once the teacher is aligned using this procedure, we distill its knowledge into the student using standard distillation techniques. The resulting teacher provides concise, pedagogically valuable supervision, leading to improved student performance.

\section{Experiments}
We conduct experiments to answer five research questions: \textbf{RQ1:} How does TeacherGRPO perform across different distillation methods? \textbf{RQ2:} What is the contribution of each component in TeacherGRPO? \textbf{RQ3:} How does TeacherGRPO compare to alternative paradigms for addressing the Gap Curse? \textbf{RQ4:} How does CSA improve alignment effectiveness? \textbf{RQ5:} How does IALR balance alignment and reasoning quality?

\subsection{Experimental Setup}
Unless otherwise stated, we evaluate TeacherGRPO across four reasoning benchmarks: Date Understanding (Date)~\citep{srivastava2023beyond}, StrategyQA (SQA)~\citep{geva2021did}, ARC-Challenge (ARC)~\citep{clark2018think}, and CommonsenseQA (CQA)~\citep{talmor2019commonsenseqa}. We experiment with two model families: Qwen (teacher: Qwen2.5-3B-Instruct, student: Qwen2.5-0.5B-Instruct) and Gemma (teacher: Gemma-3-1b-it, student: Gemma-3-270m-it). We benchmark our approach against baselines without Teacher Alignment and with KD-based alignment, alongside several standard distillation methods: Knowledge Distillation (KD)~\citep{hinton2015distilling}, Sequence-Level KD (SeqKD)~\citep{kim2016sequence}, and Adaptive Off-Policy Distillation~\citep{ko2024distillm}. Comprehensive implementation details are provided in Appendix~\ref{sec:appendix_setup}.

\subsection{Experiment Results and Analyses}

\noindent\textbf{\textit{Main Results.}} To address \textbf{RQ1}, we compare TeacherGRPO against vanilla distillation and KD-based Teacher Alignment (TeacherKD) across four reasoning benchmarks and four distillation methods using Qwen and Gemma model families. We have three key observations from Table~\ref{tab:main_results}:

(1) \textbf{TeacherGRPO consistently demonstrates robust performance gains.} For the Qwen family, TeacherGRPO achieves average improvements of 7.4 points over vanilla distillation methods, while the improvements average 3.6 points for the Gemma family. These consistent gains demonstrate TeacherGRPO's effectiveness at aligning teacher reasoning to student capacity, enabling smaller models to achieve substantially improved reasoning performance.

(2) \textbf{TeacherKD suffers from unstable performance.} While TeacherKD occasionally achieves strong results, where on Gemma with KD where it reaches 47.6 on StrategyQA (even 3.1 points above TeacherGRPO), it catastrophically fails in other settings. For instance, on Qwen with KD distillation, TeacherKD drops 20.5 point from the vanilla baseline of 47.3. This instability stems from the collapse phenomenon identified in Section~\ref{sec:preliminary}.

(3) \textbf{TeacherGRPO preserves teacher reasoning capabilities while TeacherKD degrades them.} TeacherKD-aligned teachers consistently underperform the base teacher: Qwen accuracy drops 5.4 points, and Gemma drops 2.3 points. In contrast, TeacherGRPO-aligned teachers maintain or even improve performance. This preservation ensures the aligned teacher remains a high-quality knowledge source throughout distillation.

\begin{table}[t]
\centering
\caption{Ablation results on Date and ARC with Qwen. Best values are in bold.}
\label{tab:ablation_study}
\resizebox{0.85\linewidth}{!}{
\renewcommand{\arraystretch}{0.86}
\begin{tabular}{lccc}
\toprule
\textbf{Method} & \textbf{Date} & \textbf{ARC} & \textbf{Avg} \\
\midrule
\multicolumn{4}{l}{KD} \\
\textbf{TeacherGRPO} & 40.2 & 45.2 & \textbf{42.7} \\
\hspace{3mm}\textit{w/o CSA} & 34.9 & 43.6 & 39.3 \\
\hspace{3mm}\textit{w/o IALR} & 36.4 & 42.8 & 39.6 \\
\midrule
\multicolumn{4}{l}{SeqKD} \\
\textbf{TeacherGRPO} & 38.8 & 41.3 & \textbf{40.1} \\
\hspace{3mm}\textit{w/o CSA} & 36.7 & 37.8 & 37.3 \\
\hspace{3mm}\textit{w/o IALR} & 35.9 & 34.8 & 35.4 \\
\midrule
\multicolumn{4}{l}{Adaptive Off-Policy} \\
\textbf{TeacherGRPO} & 30.2 & 44.6 & 37.4 \\
\hspace{3mm}\textit{w/o CSA} & 28.7 & 44.1 & 36.4 \\
\hspace{3mm}\textit{w/o IALR} & 29.6 & 46.1 & \textbf{37.9} \\
\bottomrule
\end{tabular}
}
\end{table}

\noindent\textbf{\textit{Ablation Study.}} To address \textbf{RQ2}, we systematically ablate CSA and IALR components and evaluate their individual contributions on Date Understanding and ARC-Challenge using the Qwen model family. We observe in Table~\ref{tab:ablation_study} that both components are essential for TeacherGRPO's performance, where ablating CSA decreases performance by 2.4 on average, and removing IALR leads to drops of 2.4 on average. Notably, the impact of each component varies by distillation method: CSA proves more critical for logit-based methods (KD) where distribution-level alignment is paramount, while IALR shows greater importance for SeqKD where sequence length affects training data quality.

\begin{table}[t]
\centering
\caption{Performance comparison of different paradigms on Date and ARC with Qwen.}
\label{tab:paradigm}
\resizebox{0.95\linewidth}{!}{
\renewcommand{\arraystretch}{1.05}
\begin{tabular}{lccc}
\toprule
\textbf{Method} & \textbf{Date} & \textbf{ARC} & \textbf{Avg} \\
\midrule
KD &  &  &  \\
\hspace{3mm}\textit{+ Data-Centric Selection} & 32.5 & 43.9 & 38.2 \\
\hspace{3mm}\textit{+ Intermediate Supervision} & 34.9 & 41.2 & 38.1 \\
\hspace{3mm}\textit{+ \textbf{TeacherGRPO}} & 40.2 & 45.2 & \textbf{42.7} \\
\midrule
SeqKD &  &  &  \\
\hspace{3mm}\textit{+ Data-Centric Selection} & 36.1 & 40.6 & 38.4 \\
\hspace{3mm}\textit{+ Intermediate Supervision} & 36.7 & 41.9 & 39.3 \\
\hspace{3mm}\textit{+ \textbf{TeacherGRPO}} & 38.8 & 41.3 & \textbf{40.1} \\
\midrule
Adaptive Off-Policy &  &  &  \\
\hspace{3mm}\textit{+ Data-Centric Selection} & 26.6 & 42.2 & 34.4 \\
\hspace{3mm}\textit{+ Intermediate Supervision} & 27.2 & 41.8 & 34.5 \\
\hspace{3mm}\textit{+ \textbf{TeacherGRPO}} & 30.2 & 44.6 & \textbf{37.4} \\
\bottomrule
\end{tabular}
}
\end{table}

\noindent\textbf{\textit{Paradigm Comparison.}} To address \textbf{RQ3}, we compare TeacherGRPO against two alternative paradigms for addressing the Gap Curse on Date Understanding and ARC-Challenge using the Qwen family. For Data-Centric Selection, we filter out 10\% the most challenging examples~\citep{liu2025less}. For Intermediate Supervision, we introduce an assistant teacher (Qwen2.5-1.5B)~\citep{zhou2024teaching}. From the results in Table~\ref{tab:paradigm}, we observe that TeacherGRPO consistently outperforms both alternatives, with average improvements of 4.3 points over Data-Centric Selection and 3.4 points over Intermediate Supervision. These results demonstrate that well-designed Teacher Alignment without discarding valuable training examples or relying on weaker intermediate models, provides a more effective solution to the Gap Curse.

\begin{figure}[t]
    \centering
    \includegraphics[width=\linewidth]{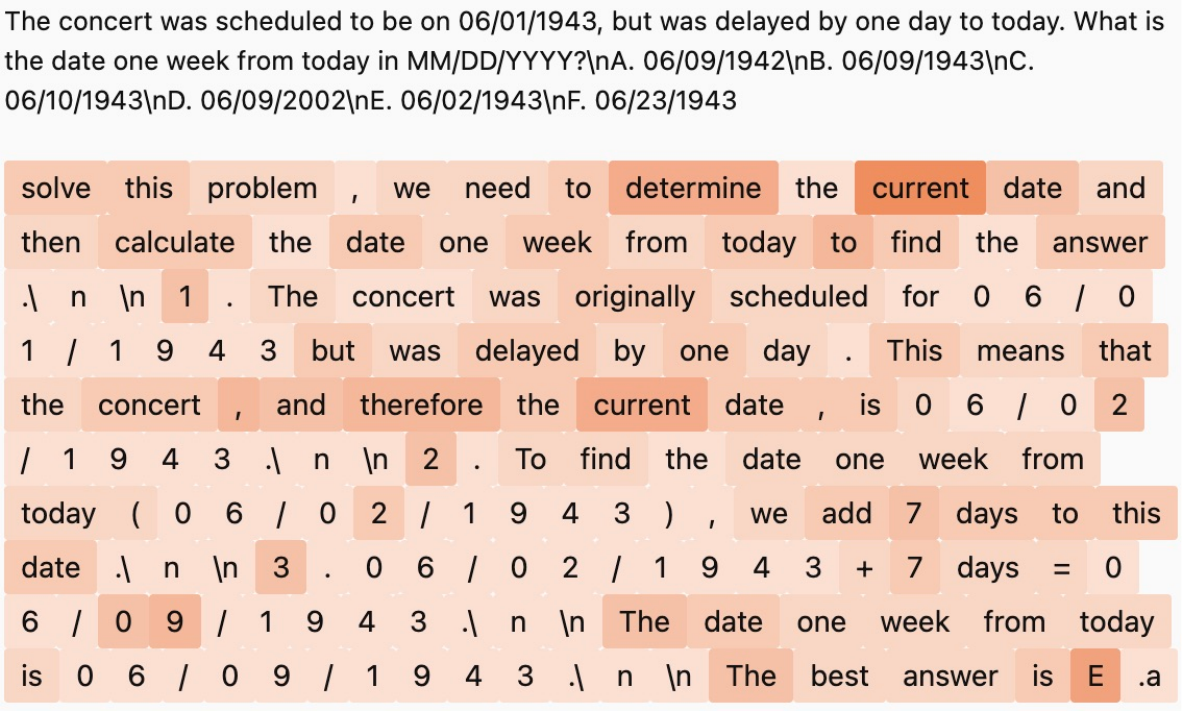}
    \caption{\textbf{KL divergence visualization for a sample path.} Darker tokens correspond to high-divergence.}
    \label{fig:gap}
\end{figure}

\noindent\textbf{\textit{CSA Analysis.}} To address \textbf{RQ4}, we compare TeacherGRPO's curriculum-based alignment against static strategies that use only the top 30\% or bottom 30\% of tokens and distribution entries ranked by teacher-student divergence. Table~\ref{tab:curriculum} presents results on Date Understanding with Qwen using KD distillation. The curriculum approach substantially outperforms both static strategies, demonstrating its effectiveness at progressively reducing noise while preserving information. Notably, the bottom 30\% performs dramatically worse, confirming that low-divergence regions hardly contain beneficial alignment signals.
To illustrate CSA's mechanism, Figure~\ref{fig:gap} visualizes token-level KL divergence for a sample reasoning path. We observe that high-divergence tokens consistently correspond to reasoning-critical elements. For example, ``current'' marks a pivotal intermediate state, while connectives like ``but'' and ``and'' signal logical transitions between reasoning steps.

\begin{table}[t]
\centering
\caption{Student performance using top 30\%, bottom 30\%, or curriculum-based token and distribution selection on Date Understanding with Qwen.}
\label{tab:curriculum}
\renewcommand{\arraystretch}{0.86}
\begin{tabular}{lcc}
\toprule
\textbf{Method} & \textbf{Token} & \textbf{Distribution}\\
\midrule
\textbf{Curriculum} & 40.2 & 40.2  \\
\textbf{Top 30\%} & 36.1 & 35.5  \\
\textbf{Bottom 30\%} & 30.2 & 27.2  \\
\bottomrule
\end{tabular}
\end{table}

\begin{figure}[t]
  \centering
  \includegraphics[width=\linewidth]{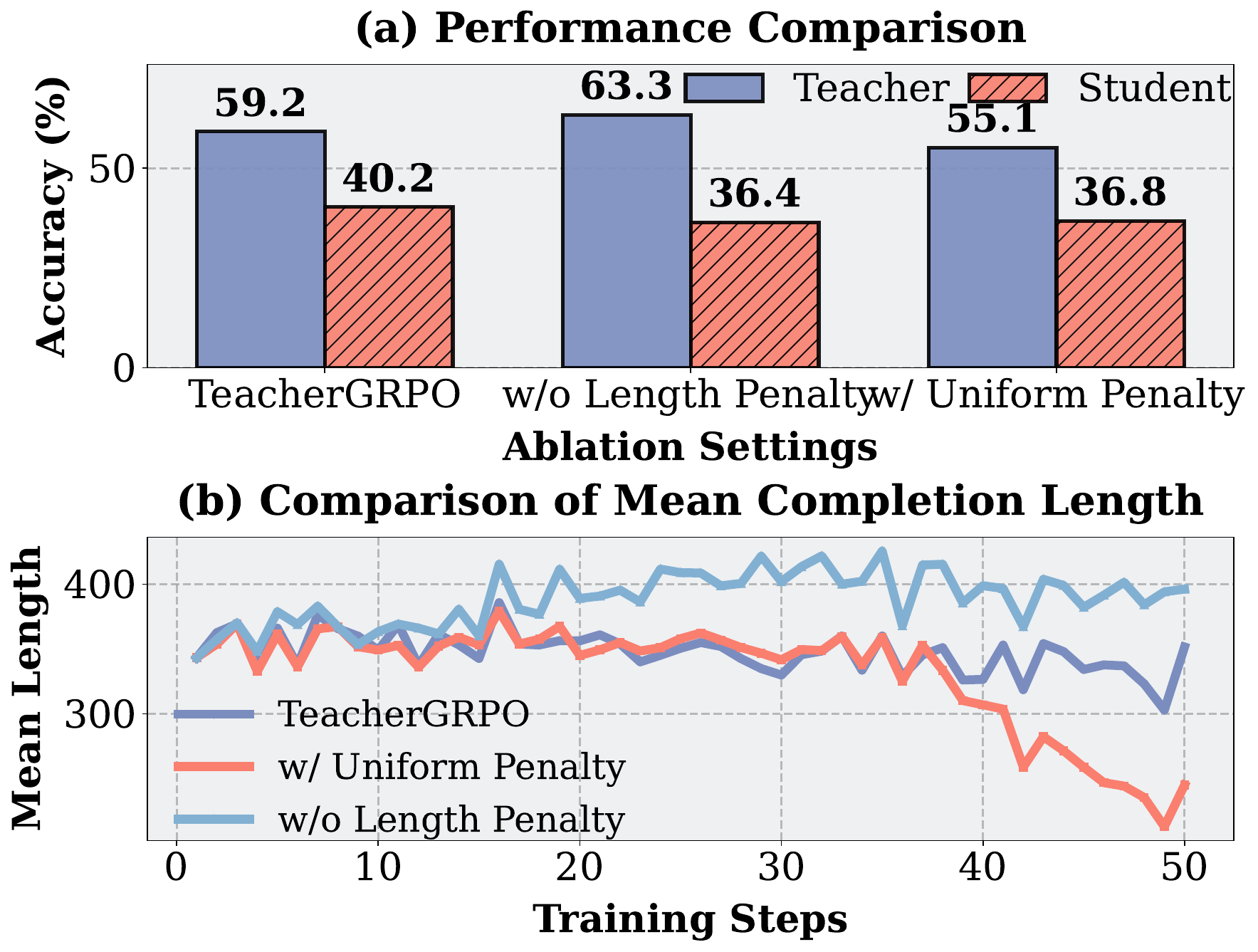}
  \caption{\textbf{Impact of length penalty.} (a) Student performance and (b) generation length during Alignment.}
  \label{fig:length_penalty}
  \vspace{-5pt}
\end{figure}

\noindent\textbf{\textit{IALR Analysis.}} To address \textbf{RQ5}, we compare TeacherGRPO's importance-adaptive length regularization against two baselines: uniform length penalty and no length penalty. We track both student performance and mean generation length during Teacher Alignment on Date Understanding with Qwen using KD distillation. Figure~\ref{fig:length_penalty}(a) shows that TeacherGRPO outperforms both uniform penalty and no penalty, demonstrating IALR's effectiveness at preventing length exploitation without eliminating essential reasoning steps. The alignment dynamics in Figure~\ref{fig:length_penalty}(b) provide further validation, where IALR achieves gradual, controlled reduction, while no penalty increases verbosity and uniform penalty causes abrupt compression.

\section{Related Work}

\noindent\textbf{Gap Curse in Knowledge Distillation.}
Knowledge distillation~\citep{hinton2015distilling} transfers capabilities from large language models to smaller ones by minimizing KL divergence between output distributions~\citep{xu2024survey}. Recent advances include white-box methods such as On-Policy Distillation~\citep{agarwal2024policy} and Adaptive Off-Policy Distillation~\citep{ko2024distillm}, alongside black-box methods like SeqKD~\citep{kim2016sequence}. Despite these innovations, distillation methods fail to escape the Gap Curse: increasingly powerful teachers degrade rather than improve student performance~\citep{zhang2025towards, zhu2022teach, jiang2025drp}, stemming from capacity mismatch~\citep{ji2025chain}. Current solutions employ either \textit{Data-Centric Selection}, utilizing easier examples~\citep{liu2025less, huang2025selectkd, li2025sample, just2025distilling}, or \textit{Intermediate Supervision}, introducing assistant models~\citep{zhou2024teaching, lv2024taekd, ding2025micota}. Both paradigms sacrifice supervision breadth or quality~\citep{budiman2025survey}.

\noindent\textbf{Teacher-Side Solution for Gap Curse.}
In vision tasks, an alternative perspective treats the gap curse as a teacher-side problem requiring adaptation. Methods such as reverse knowledge distillation~\citep{nasser2024reverse}, pruning~\citep{park2022prune}, and student-friendly distillation~\citep{yuan2024student} create ``easier'' teachers in vision tasks. However, these methods are not suitable for LLM reasoning distillation as they risk catastrophic collapse~\citep{li2024continual}, degrading logical correctness as teachers overfit to student errors. In this work, TeacherGRPO reframes Teacher Alignment as an RL problem with specialized rewards that preserve pedagogical value, reviving alignment-based methods for reasoning distillation.

\section{Conclusion}

We introduced Teacher Alignment, a paradigm that adapts teachers to match student capacity rather than discarding data or relying on weaker intermediaries. To address the catastrophic collapse of naive alignment methods, we presented TeacherGRPO, a reinforcement learning framework built on GRPO with two key innovations: CSA filters token- and distribution-level noise to focus on high-signal reasoning gaps, while IALR penalizes verbosity without truncating essential reasoning. Extensive experiments across diverse benchmarks and distillation methods demonstrate the effectiveness of TeacherGRPO. Our work establishes Teacher Alignment as an effective solution to the Gap Curse, enabling more effective knowledge transfer from powerful reasoning models.

\section*{Acknowledgment}
This work was supported in part by the National Science Foundation (NSF) under Grants IIS-2144209, IIS-2223769, BCS-2228534, CMMI-2411248, and IIBR-2601942; by the Office of Naval Research (ONR) under Grant N000142412636; and by the Commonwealth Cyber Initiative (CCI) under Grant HV-4Q26-073.

\section*{Limitations}

\noindent\textbf{Model Scale.} Our work focuses on the small-model 
reasoning distillation regime, where the Gap Curse is most pronounced 
and efficient deployment matters most in practice. Within this scope, 
we validate TeacherGRPO on two distinct model families (Qwen and Gemma) 
with different architectures and compression ratios (6$\times$ and 
3.7$\times$), providing evidence of cross-architecture generalizability. 
That said, extending experiments to larger teacher-student pairs 
(e.g., 7B$\to$1.5B or beyond) would further characterize how 
TeacherGRPO scales with model capacity, which we leave to future work 
given current computational constraints.

\noindent\textbf{Task Coverage.} Following standard evaluation 
protocols in recent reasoning distillation 
work~\citep{liu2025less, zhou2024teaching, ko2024distillm}, we 
evaluate on four widely-adopted benchmarks spanning temporal, 
multi-hop, scientific, and commonsense reasoning. The Gap Curse phenomenon we address is most clearly characterized in 
tasks with verifiable answers, making our chosen benchmarks an 
appropriate testbed for the core contribution. However, broader task 
coverage remains a natural direction for future work.

\noindent\textbf{Computational Overhead.} TeacherGRPO introduces an 
additional alignment phase before standard distillation. However, this cost 
is amortized across all downstream students distilled from the same 
aligned teacher, while our LoRA-based implementation keeps the overhead 
modest. Reducing the overhead 
through more efficient alignment objectives is an interesting direction 
for further optimization and can make it more practical.

\section*{Use of AI Assistants}
We mainly used AI writing assistants to polish the phrasing and grammar of this manuscript. We take full responsibility for all content in this paper.

\bibliography{custom}

\newpage
\appendix

\section{Extended Experimental Results}
\label{app:extended_results}

This section provides additional experiments that further probe TeacherGRPO's generality across task types, distillation algorithms, and hyperparameter settings.

\subsection{Evaluation on Mathematical Reasoning}
\label{app:math}

The four benchmarks in the main paper (Date, SQA, ARC, CQA) are all structured as multiple-choice or binary classification tasks. To examine whether TeacherGRPO's benefits extend to open-ended, multi-step reasoning, we evaluate on \textbf{MATH}~\citep{hendrycks2021measuring}, a benchmark of competition-level mathematical problems requiring long, intricately dependent reasoning chains. Table~\ref{tab:math} reports accuracy for both model families across all three distillation pipelines.

TeacherGRPO consistently improves over vanilla distillation on both model families, while TeacherKD is unstable and occasionally underperforms the vanilla baseline (e.g., KD+TeacherKD on Qwen drops from 24.8 to 22.1). We note that the small students (0.5B and 270M parameters) have limited base capacity for competition-level math, particularly Gemma-3-270M, which compresses the achievable improvement and makes TeacherGRPO's absolute gains smaller than on the other benchmarks. Nevertheless, the consistent direction of improvement suggests that the delimiter-based step segmentation and entropy-based importance estimation underlying IALR remain effective even when reasoning steps are more tightly interdependent than in the multiple-choice benchmarks.

\begin{table}[h]
\centering
\resizebox{\columnwidth}{!}{%
\begin{tabular}{lcc}
\toprule
\textbf{Method} & \textbf{Qwen2.5 (3B$\to$0.5B)} & \textbf{Gemma3 (1B$\to$270M)} \\
\midrule
Base & 20.9 & 3.7 \\
\midrule
KD & 24.8 & 4.1 \\
\hspace{3mm}\textit{+ TeacherKD} & 22.1 & 3.9 \\
\hspace{3mm}\textit{+ \textbf{TeacherGRPO}} & \textbf{25.5} & \textbf{4.2} \\
\midrule
SeqKD & 25.6 & 3.5 \\
\hspace{3mm}\textit{+ TeacherKD} & \textbf{30.0} & 3.8 \\
\hspace{3mm}\textit{+ \textbf{TeacherGRPO}} & 28.1 & \textbf{4.1} \\
\midrule
Adaptive Off-Policy & 28.2 & 3.9 \\
\hspace{3mm}\textit{+ TeacherKD} & 24.0 & 3.5 \\
\hspace{3mm}\textit{+ \textbf{TeacherGRPO}} & \textbf{28.8} & \textbf{4.4} \\
\bottomrule
\end{tabular}%
}
\caption{Accuracy on \textbf{MATH}, a multi-step mathematical reasoning benchmark with longer and more interdependent reasoning chains than the main benchmarks. TeacherGRPO consistently improves over vanilla distillation across both model families and all three distillation pipelines, while TeacherKD remains unstable.}
\label{tab:math}
\end{table}

\subsection{Comparison with a Stronger Distillation Baseline}
\label{app:speculative}

To verify that TeacherGRPO's benefits are not confined to the classic distillation algorithms evaluated in the main results, we apply Teacher Alignment on top of \textbf{Speculative Knowledge Distillation (SpeculativeKD)}~\citep{xu2025speculative}, a more recent distillation method that bridges the teacher-student gap through interleaved sampling. We use the Qwen family (teacher: Qwen2.5-3B-Instruct, student: Qwen2.5-0.5B-Instruct) and evaluate on all four main benchmarks.

As shown in Table~\ref{tab:speculative}, TeacherGRPO consistently improves SpeculativeKD across all four benchmarks, whereas TeacherKD again shows instability (e.g., dropping ARC from 39.68 to 35.58 and CommonsenseQA from 47.42 to 39.23). This indicates that Teacher Alignment is complementary to advances in the underlying distillation algorithm rather than tied to any specific pipeline, and that TeacherGRPO's stability advantage over naive KD-based alignment holds beyond the three pipelines studied in the main paper.

\begin{table}[h]
\centering
\resizebox{\columnwidth}{!}{%
\begin{tabular}{lcccc}
\toprule
\textbf{Method} & \textbf{Date} & \textbf{SQA} & \textbf{ARC} & \textbf{CQA} \\
\midrule
Base (Qwen2.5-0.5B) & 20.12 & 35.81 & 37.37 & 39.14 \\
\midrule
SpeculativeKD & 24.26 & 44.98 & 39.68 & 47.42 \\
\hspace{3mm}\textit{+ TeacherKD} & 22.49 & \textbf{50.22} & 35.58 & 39.23 \\
\hspace{3mm}\textit{+ \textbf{TeacherGRPO}} & \textbf{25.44} & 50.66 & \textbf{41.72} & \textbf{47.50} \\
\bottomrule
\end{tabular}%
}
\caption{Results on SpeculativeKD~\citep{xu2025speculative} with the Qwen model family. TeacherGRPO improves SpeculativeKD across all four benchmarks, while TeacherKD is unstable and degrades performance on ARC and CommonsenseQA.}
\label{tab:speculative}
\end{table}

\subsection{Hyperparameter Sensitivity Analysis}
\label{app:sensitivity}

TeacherGRPO's reward function introduces several hyperparameters ($k_{\text{min}}$, $k_{\text{max}}$ for the token-level curriculum, $p_{\text{min}}$, $p_{\text{max}}$ for the distribution-level curriculum, and $\eta$, $\beta$ for IALR). To assess whether TeacherGRPO's performance is brittle with respect to these choices, we sweep the IALR penalty strength $\beta$, the most directly tunable of these hyperparameters, over an order of magnitude while holding all other hyperparameters at their default values. We evaluate on Date Understanding with the Qwen family using KD distillation.

Table~\ref{tab:sensitivity} shows that performance is stable within a reasonable neighborhood of the default value ($\beta \in [0.005, 0.02]$ varies by at most 3.2 points) and degrades gracefully rather than collapsing at the order-of-magnitude extremes ($\beta = 0.001$ or $\beta = 0.05$). This indicates that TeacherGRPO does not require precise tuning of $\beta$ to achieve strong performance, supporting the practicality of the overall pipeline.

\begin{table}[h]
\centering
\begin{tabular}{lccccc}
\toprule
\textbf{$\beta$} & 0.001 & 0.005 & \textbf{0.01} & 0.02 & 0.05 \\
\midrule
\textbf{Accuracy} & 36.4 & 37.0 & \textbf{40.2} & 39.4 & 35.8 \\
\bottomrule
\end{tabular}
\caption{Sensitivity of student accuracy to the IALR penalty strength $\beta$ on Date Understanding with Qwen (KD distillation). The default value ($\beta=0.01$, bolded) performs best, and performance degrades gracefully away from it rather than collapsing.}
\label{tab:sensitivity}
\end{table}

\section{Case Study for IALR}

To illustrate the impact of IALR, we present a qualitative comparison of teacher generations across four configurations: (1) \textbf{Vanilla} (no teacher alignment), (2) \textbf{TeacherGRPO}, (3) \textbf{Uniform Penalty}, and (4) \textbf{w/o IALR}.

\begin{figure}[t]
    \centering
    \includegraphics[width=\linewidth]{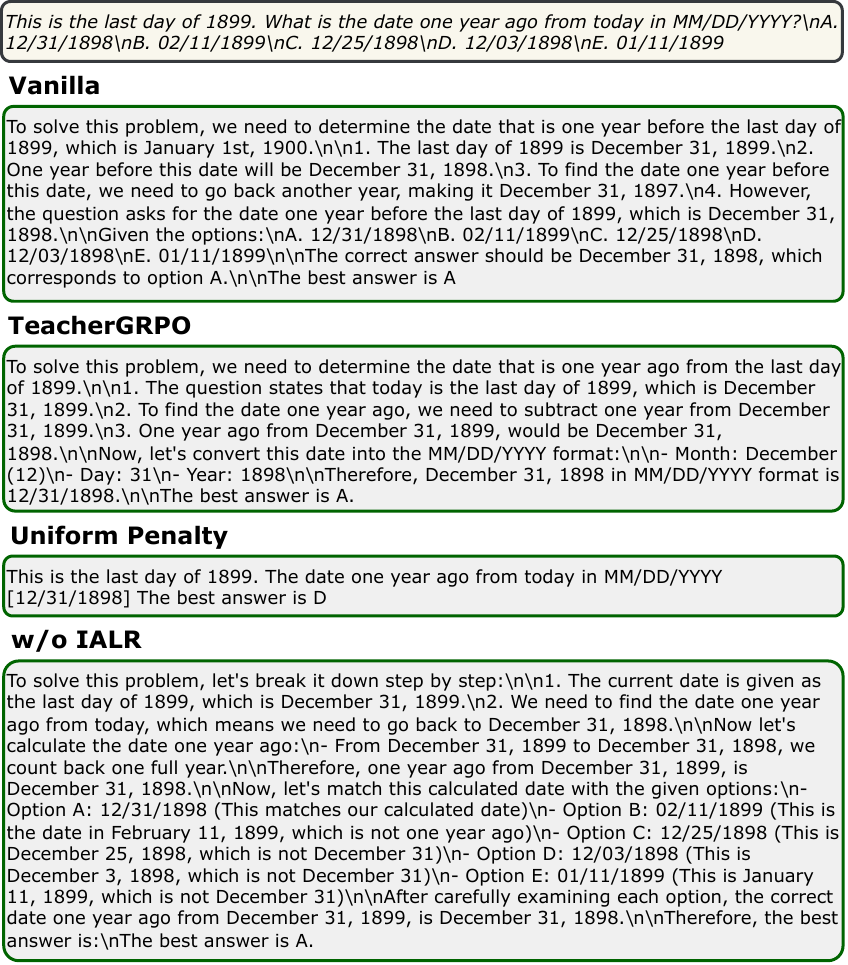}
    \caption{\textbf{Qualitative comparison of teacher generations with different length penalties.} We compare reasoning outputs from four teacher configurations on a Date Understanding example: Vanilla (no alignment), TeacherGRPO, Uniform Penalty, and w/o IALR.}
    \label{fig:case}
\end{figure}

As shown in Figure~\ref{fig:case}, \textbf{TeacherGRPO}'s generation demonstrates optimal balance between conciseness and reasoning completeness. Compared to the vanilla teacher, TeacherGRPO produces responses that are: (1) more structured, explicitly formatting the answer for clarity (e.g. ``convert this date into the MM/DD/YYYY format''), and (2) more direct in their reasoning chain (e.g., ``subtract one year'' versus ``go back another year''). These characteristics make the reasoning path more accessible for student learning while preserving logical correctness.
In contrast, the \textbf{Uniform Penalty} approach produces overly compressed outputs that sacrifice essential reasoning steps. This aggressive truncation eliminates critical intermediate logic, ultimately leading to an incorrect answer.
The \textbf{w/o IALR} variant exhibits the opposite problem: excessive verbosity without pedagogical value. While it systematically examines each answer choice with detailed explanations, the resulting response is unnecessarily long and difficult for capacity-constrained students to process effectively. This demonstrates that length regularization without importance weighting fails to distinguish between essential reasoning steps and redundant elaboration.

These examples validate IALR's core mechanism: by adaptively penalizing based on step importance, TeacherGRPO produces reasoning that are concise yet complete, optimizing for student digestibility without compromising logical integrity.

\section{Training Stability Analysis}

\begin{figure}[t]
    \centering
    \includegraphics[width=\linewidth]{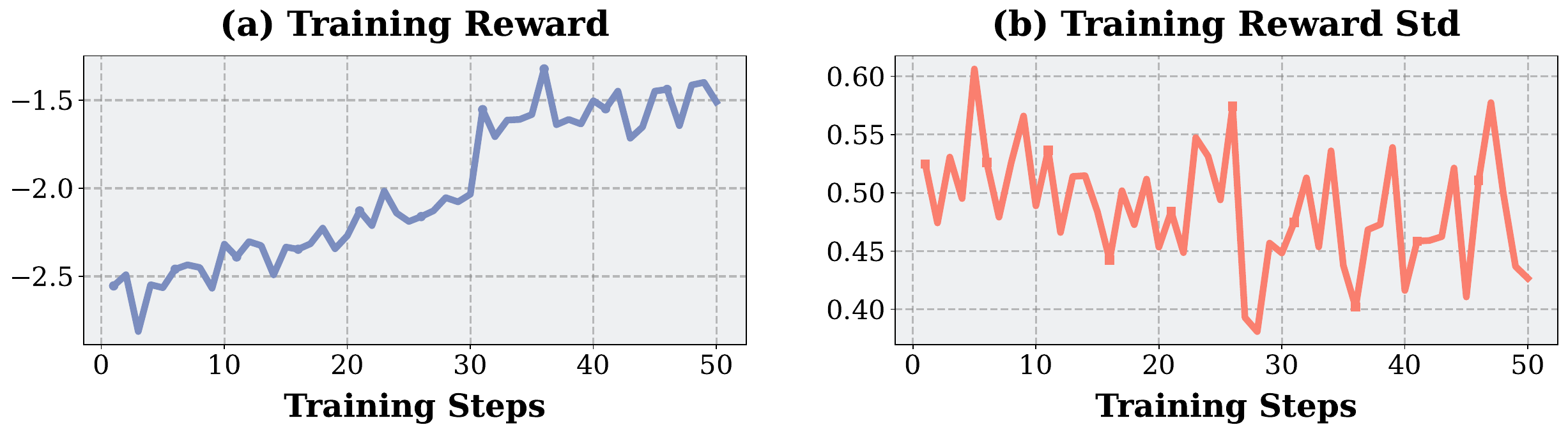}
    \caption{\textbf{Training dynamics during Teacher Alignment on the Date Understanding dataset.} Both curves are smoothed using exponential moving average (EMA weight = 0.9).}
    \label{fig:reward_analysis}
\end{figure}

To evaluate the optimization stability of TeacherGRPO, we analyze the training dynamics during the Teacher Alignment phase on the Date Understanding dataset. Figure~\ref{fig:reward_analysis} presents the training reward (left) and its standard deviation (right), both smoothed using exponential moving average with weight 0.9.

We observe two key indicators of stable convergence: (1) \textbf{Steady reward improvement}: The mean reward increases consistently from approximately -2.6 to -1.5 over 50 training steps, demonstrating effective optimization toward the alignment objective. (2) \textbf{Decreasing reward variance}: The reward standard deviation exhibits a gradual declining trend from around 0.60 to 0.45, indicating that the policy becomes increasingly confident and consistent in its generations.

These training dynamics confirm that TeacherGRPO achieves stable optimization and validates the effectiveness of CSA in maintaining stable learning signals throughout the alignment process.

\section{Experiment Setup Details}
\label{sec:appendix_setup}

\subsection{Datasets}
We evaluate our approach across four diverse reasoning benchmarks that require different types of reasoning capabilities:
\begin{itemize}
    \item \textbf{Date Understanding (Date)}~\citep{srivastava2023beyond}: Temporal reasoning tasks requiring models to understand and manipulate date-related information.
    \item \textbf{StrategyQA (SQA)}~\citep{geva2021did}: Multi-hop reasoning questions requiring implicit strategy decomposition and commonsense knowledge integration.
    \item \textbf{ARC-Challenge (ARC)}~\citep{clark2018think}: Science questions from standardized tests requiring commonsense and scientific reasoning.
    \item \textbf{CommonsenseQA (CQA)}~\citep{talmor2019commonsenseqa}: Multiple-choice questions testing commonsense knowledge and reasoning capabilities across diverse everyday scenarios.
\end{itemize}

\subsection{Model Selection}
We conduct experiments on two distinct model families to demonstrate the generalizability of our approach across different architectures and compression ratios:
\begin{itemize}
    \item \textbf{Qwen Family}~\citep{yang2025qwen3}: Teacher model is \texttt{Qwen2.5-3B-Instruct} (3B parameters), and student model is \texttt{Qwen2.5-0.5B-Instruct} (0.5B parameters), yielding a 6$\times$ compression ratio.
    \item \textbf{Gemma Family}~\citep{team2025gemma}: Teacher model is \texttt{Gemma-3-1b-it} (1B parameters), and student model is \texttt{Gemma-3-270m-it} (270M parameters), achieving approximately 3.7$\times$ compression.
\end{itemize}
Both model families represent state-of-the-art open-source instruction-tuned models with strong baseline reasoning capabilities.

\subsection{Distillation Methods}
We apply TeacherGRPO to several state-of-the-art distillation approaches to demonstrate its compatibility and effectiveness across different distillation paradigms:
\begin{itemize}
    \item \textbf{Knowledge Distillation (KD)}~\citep{hinton2015distilling}: Logit-based knowledge distillation using responses from Gemini 2.5 Pro as ground truth labels. The student minimizes KL divergence between its output distribution and the teacher's softened probability distribution.
    \item \textbf{Sequence-Level KD (SeqKD)}~\citep{kim2016sequence}: Token-level distillation where the teacher generates training sequences via sampling, and the student learns to reproduce these sequences through standard cross-entropy training.
    \item \textbf{Adaptive Off-Policy Distillation}~\citep{ko2024distillm}: An efficient variant of iterative on-policy distillation~\citep{agarwal2024policy} where the student first generates responses, which are then stored and replayed during training. This approach maintains effectiveness while significantly reducing computational overhead compared to standard on-policy methods.
\end{itemize}

\subsection{Baseline Methods}
We compare TeacherGRPO against the following baselines:
\begin{itemize}
    \item \textbf{Vanilla Distillation}: Distillation without any teacher alignment, serving as the primary baseline.
    \item \textbf{TeacherKD}: Our implementation of knowledge distillation for teacher alignment, where the teacher is adapted toward the student's distribution via standard KL divergence minimization.
    \item \textbf{Data-Centric Selection}~\citep{liu2025less}: Filters out the 10\% most challenging examples based on teacher-student divergence metrics, representing the data filtering paradigm.
    \item \textbf{Intermediate Supervision}~\citep{zhou2024teaching}: Introduces an intermediate assistant teacher (\texttt{Qwen2.5-1.5B-Instruct} for the Qwen family) positioned between the teacher and student to bridge the capability gap.
\end{itemize}

\subsection{Implementation Details}

\subsubsection{Teacher Alignment Phase}
\textit{TeacherKD.} We train the teacher to minimize KL divergence with the student using a learning rate of 5$\times$10$^{-6}$, batch size 4, and AdamW optimizer with $\beta_1=0.9$, $\beta_2=0.999$. We train for one complete epoch on Date Understanding and limit training to 100 steps maximum on other datasets to prevent catastrophic collapse.

\textit{TeacherGRPO.} We employ GRPO with a learning rate of 2$\times$10$^{-5}$, batch size 4, gradient accumulation over 32 steps, and sample 4 generations per query for group-based optimization. We use AdamW optimizer with the same $\beta$ values. We train for five epochs on Date Understanding and limit training to 100 steps maximum on other datasets. The GRPO-specific hyperparameters are set as follows: clip ratio $\epsilon_{clip}=0.2$, KL penalty coefficient $\gamma=0.01$, and group size $G=4$.

All teacher alignment is performed using LoRA~\citep{hu2022lora} with rank $r=32$, $\alpha=64$, and dropout $p=0.1$ for memory efficiency, adapting only the attention projection matrices (query, key, value, and output).

\subsubsection{Student Distillation Phase}
Following teacher alignment, we distill knowledge into student models using a learning rate of 2$\times$10$^{-5}$, batch size 4, and AdamW optimizer. We train for two epochs on Date Understanding and ARC-Challenge, and one epoch on StrategyQA and CommonsenseQA. For KD-based distillation, we use a temperature of $T=2.0$ and balance coefficient $\alpha=0.5$ between the distillation loss and task loss. All student distillation is performed with full fine-tuning.

\subsubsection{TeacherGRPO Hyperparameters}
For Curriculum Selective Alignment (CSA), we set token-level curriculum bounds to $k_{\text{min}}=0.3$ and $k_{\text{max}}=1.0$, and distribution-level curriculum bounds to $p_{\text{min}}=0.3$ and $p_{\text{max}}=1.0$. For Importance-Adaptive Length Regularization (IALR), we use temperature $\eta=4$ for importance score computation and penalty strength $\beta=0.01$. The reward combination weights are set to $\alpha_1=0.6$ for CSA reward, $\alpha_2=0.6$ for IALR reward, and $\alpha_3=0.4$ for verification reward.

\subsubsection{Evaluation Protocol}
We evaluate all models using greedy decoding with maximum generation length of 1024 tokens. For multiple-choice questions (ARC-Challenge, CommonsenseQA, Date Understanding), we extract the model's final answer by pattern matching on the response format specified in the prompt. For StrategyQA, we extract binary True/False predictions. We report accuracy as the primary metric across all datasets.

\subsubsection{Infrastructure}
Our implementation is based on PyTorch 2.1.0 with Python 3.11.7. We use the Transformers library (version 4.36.0) for model loading and the TRL library (version 0.7.4) for GRPO implementation. All experiments are conducted on NVIDIA A100 GPUs with 80GB memory.

\subsection{Prompts}
We provide the exact prompts used for generation across all datasets below.

\begin{promptbox}{{Prompt for ARC-Challenge, Date, and CommonsenseQA}}
Given the following question and \{num\_choices\} candidate answers (\{choice\_str\}), choose the best answer.

Question: \{question\}

\{formatted\_choices\}

Please reason step by step, and conclude with your choice. Your response should end with ``The best answer is []'' where the [] is one of \{choice\_str\}.
\end{promptbox}

\begin{promptbox}{Prompt for StrategyQA}
Question: \{question\}

Please reason step by step, and conclude with either ``True'' or ``False''.
\end{promptbox}

\end{document}